\documentclass{article}

\usepackage{PRIMEarxiv}

\usepackage[utf8]{inputenc} 
\usepackage[T1]{fontenc}    
\usepackage{hyperref}       
\usepackage{url}            
\usepackage{booktabs}       
\usepackage{amsfonts}       
\usepackage{nicefrac}       
\usepackage{microtype}      
\usepackage{lipsum}
\usepackage{fancyhdr}       
\usepackage{graphicx}       
\usepackage{multirow}
\graphicspath{{media/}}     

\title{LegalTurk Optimized BERT for Multi-Label Text Classification and NER
}

\author{
  Farnaz Zeidi\thanks{Corresponding author} \\
  Department of Informatics \\
  Istanbul University \\
  Istanbul, Turkey \\
  \texttt{farnaz.zeidi@ogr.iu.edu.tr} \\
  \And
  Mehmet Fatih Amasyali \\
  Department of Computer Engineering \\
  Yildiz Technical University \\
  Istanbul, Turkey \\
  \texttt{amasyali@yildiz.edu.tr} \\
  \And
  Çiğdem Erol \\
  Faculty of Science, Department of Computer Sciences \\
  Istanbul University \\
  Istanbul, Turkey \\
  \texttt{cigdems@istanbul.edu.tr} \\
}

\begin{document}
\maketitle

\begin{abstract}
The introduction of the Transformer neural network, along with techniques like self-supervised pre-training and transfer learning, has paved the way for advanced models like BERT. Despite BERT's impressive performance, opportunities for further enhancement exist. To our knowledge, most efforts are focusing on improving BERT's performance in English and in general domains, with no study specifically addressing the legal Turkish domain. Our study is primarily dedicated to enhancing the BERT model within the legal Turkish domain through modifications in the pre-training phase. In this work, we introduce our innovative modified pre-training approach by combining diverse masking strategies. In the fine-tuning task, we focus on two essential downstream tasks in the legal domain: name entity recognition and multi-label text classification. To evaluate our modified pre-training approach, we fine-tuned all customized models alongside the original BERT models to compare their performance. Our modified approach demonstrated significant improvements in both NER and multi-label text classification tasks compared to the original BERT model. Finally, to showcase the impact of our proposed models, we trained our best models with different corpus sizes and compared them with BERTurk models. The experimental results demonstrate that our innovative approach, despite being pre-trained on a smaller corpus, competes with BERTurk.\end{abstract}


\section{Introduction}
The utilization of Natural Language Processing (NLP) techniques in the Legal Tech field is gaining significance, primarily due to challenges posed by lengthy legal texts and time-consuming search processes. Applications like Named Entity Recognition (NER) and text classification hold the potential to greatly enhance the performance of legal professionals by enabling quicker decision-making. The use of transformer-based models like Google's Bidirectional Encoder Representations from Transformers (BERT) \cite{devlin2018bert} within the legal domain has garnered substantial interest from numerous researchers. The state-of-the-art BERT operates through two stages: pre-training, where it learns language representations from extensive unlabeled datasets (corpora) in a self-supervised manner, followed by fine-tuning, allowing the pre-trained model to adapt to supervised downstream tasks using limited labeled data. While Google's BERT has exhibited remarkable performance, there remains an opportunity for further enhancements, particularly in customizing these models for specific domains and non-English contexts. Turkish-based models, in comparison to English-based ones, are notably limited, with most trained on general text rather than specific domains. For instance, the well-known monolingual Turkish BERT model, Schweter's BERTurk, was pre-trained using a 35GB dataset of non-specific domain Turkish text  \cite{schweter2020berturk}. In the Turkish legal domain, two recent models have emerged: BERTurk-Legal \cite{ozturk} and HukukBERT \cite{akca}. These models concentrate on further pre-training BERTurk using legal domain-specific corpora, without making alterations to the pre-training subtasks on the BERT model.
Acknowledging these circumstances, identifying weaknesses in the BERT model and aiming to enhance its performance, particularly tailored to legal Turkish text, becomes imperative. This study's scope revolves around modifying pre-training tasks in BERT, training all proposed models from scratch, and subsequently fine-tuning them to extract information from legal Turkish text. To the best of our knowledge, this is the first work focusing on a modified pre-training approach for the Turkish legal domain. The main approach of our work are summarized below:

\begin{itemize}
  \item Without making any modifications to the BERT-based configuration, we pursued three essential strategies to improve BERT's performance: replacing Next Sentence Prediction (NSP) with Sentence Order Prediction (SOP), eliminating NSP entirely, and combining Masked Language Model (MLM) with Term Frequency - Inverse Document Frequency (TF-IDF). In our innovative approach, we propose replacing 10\% of the tokens selected for MLM with those having high TF-IDF values, instead of randomly selecting tokens from the tokenizer's vocabulary. Additionally, we implemented diverse masking strategies within the original rules of MLM (80\_10\_10).
  \item To pre-train our proposed models from scratch, we utilized a 50 MB legal Turkish corpus sourced from legal-related thesis documents available in the Higher Education Board National Thesis Center (YÖKTEZ).
  \item Subsequently, these customized models were fine-tuned for NER and multi-label text classification using Turkish legal-related human-annotated datasets.
  \item We compared our proposed models with the original BERT structure. Subsequently, after identifying the best models pre-trained on 50 MB for each downstream task, in further steps, they were pre-trained on different corpus sizes (100 MB, 500 MB, 1 GB and 2 GB) and compared them with the performance of the BERTurk model.
\end{itemize}

 The remainder of the paper is organized as follows: Section 2 presents the related works on enhancing BERT's performance in languages other than Turkish and Turkish, covering legal and non-legal domains. Section 3 introduces the proposed models. Section 4 provides details about dataset collection in both the pre-training and fine-tuning phases. Section 5 outlines ways to evaluate the models' performance. Section 6 discusses the experimental results. Finally, Sections 7 and 8 include a perspective on our future work and the conclusion.

\section{Related works}
\label{sec:headings}
GPT (Generative Pre-trained Transformer) \cite{radford2018improving} and BERT \cite{devlin2018bert} were the pioneering Transformer-based pre-trained language models, with GPT based on the Transformer decoder and BERT on the Transformer encoder \cite{zhou2023comprehensive,kalyan2021ammus}. Subsequent to these, there have been advancements in models such as XLNet \cite{yang2019xlnet}, RoBERTa \cite{liu2019roberta}, ELECTRA \cite{clark2020electra}, ALBERT \cite{lan2019albert}, T5 \cite{raffel2020exploring}, BART \cite{lewis2019bart} and PEGASUS \cite{zhang2020pegasus}. Among these, XLNet, RoBERTa, ELECTRA, and ALBERT represent improvements over the BERT model, while T5, BART, and PEGASUS are encoder-decoder models. Given the focus of our study on BERT, an encoder-based model, we will specifically delve into encoder-based models that have improved BERT's performance by modifying subtasks in pre-training, including tasks like MLM and NSP. We aimed to cover studies aimed at enhancing the BERT model in languages other than Turkish and Turkish, covering both non-legal and legal domains.

\subsection{Enhancements to the BERT Model}

Regarding the studies aimed at improving the BERT results in languages other than Turkish, several models are available, such as RoBERTa (A robustly optimized BERT pre-training approach) \cite{liu2019roberta}. The authors of RoBERTa omitted the NSP task during BERT's pre-training and introduced dynamic MLM in place of the static MLM. Notably, RoBERTa made significant adjustments to key hyperparameters, including dropout rates, batch sizes, and learning rates. Additionally, it underwent pre-training with longer sequences and a larger corpus compared to BERT. SpanBERT \cite{joshi2020spanbert} also eliminates the NSP task and seeks to reconfigure the MLM task with a creative random process to select token spans instead of individual tokens, which are then masked. In addition, they introduce a novel auxiliary objective called the "span-boundary objective" to predict the complete masked span using the tokens that represent the span's boundaries. ALBERT (A Lite BERT) \cite{lan2019albert} places its emphasis on modeling inter-sentence coherence and has introduced modifications to the NSP task, leading to SOP that is specifically designed to distinguish the sequential order of sentences within a text. Also, two parameter-reduction techniques, namely "factorized embedding parameterization" and "cross-layer parameter sharing", have been implemented. Furthermore, BERTje (A Dutch BERT Model) \cite{devries2019bertje} and StructBERT \cite{wang2019structbert} also utilize SOP in the pre-training phase. In the case of StructBERT, the authors have introduced an innovative structural pre-training approach that incorporates word and sentence structures into the BERT pre-training process. Additionally, StructBERT introduces the Span Order Recovery task to enhance the MLM task. Moreover, XLNet \cite{yang2019xlnet} represents a generalized auto-regressive pre-training approach that eliminates the NSP task and replaces MLM with a permutation language modeling (PLM). In PLM, it effectively leverages the strengths of both auto-regressive and auto-encoder methods while mitigating their weaknesses. Instead of predicting tokens sequentially, PLM predicts them in a random order. XLNet also incorporates a two-stream attention mechanism and combines BERT's bidirectional capabilities with Transformer-XL's autoregressive technology. Furthermore, The authors of ELECTRA \cite{clark2020electra} omitted the NSP task and placed a stronger emphasis on the MLM task, leading to the introduction of a new concept with the "replaced token detection" (RTD) approach. Instead of masking the input, ELECTRA corrupts the input by replacing certain tokens with possible alternatives produced from a generator model.
In the examples above, attempts were made to enhance the original BERT by modifying its core structure or subtask manipulation during the pre-training stage, resulting in the creation of new models. However, an alternative approach to improving BERT models involves maintaining the BERT structure and continuous pre-training, referred to as "Further Pre-training", to customize a language model for a specific task domain \cite{bayrak2022domain} or another language \cite{eronen2023zero}. Bayrak and Issifu \cite{bayrak2022domain} aimed to achieve better results in classifying Arabic dialects by extending the pre-training of Arabic pre-trained BERT models, including AraBERT \cite{antoun2020arabert} and MARBERT \cite{abdulmageed2020arbert}, through the incorporation of additional dialectal tweets. In another instance, ABioNER \cite{boudjellal2021abioner} underwent further pre-training with an Arabic biomedical corpus, and RadBERT \cite{yan2022radbert} was subjected to additional pre-training on the English radiology report corpus, resulting in improved performance compared to the non-specific BERT-based model. When it comes to legal-focused research, it is relatively scarce compared to other domains, such as the medical field. To the best of our knowledge, in the existing literature focusing on improving BERT performance, studies have concentrated on further pre-training non-specific BERT models using small legal-related corpora. Examples of such models include German Legal BERT \cite{yeung2019effects}, Legal-BERT \cite{chalkidis2020legalbert}, Romanian jurBERT \cite{masala2021jurbert} and Legal-Vocab-BERT  \cite{khan2021comparing}.

\subsection{Turkish Language-based Enhancements to the BERT Model}
In general, there are significantly fewer studies focusing on the Turkish language compared to English-centric studies. The most prominent Turkish model is Schweter's BERTurk, a monolingual Turkish BERT model representing an original BERT model trained using a Turkish corpus by the MDZ Digital Library team. This model underwent training on a substantial dataset, encompassing 35GB of text and a total of 44,04,976,662 tokens. The nonspecific corpus was compiled from the Turkish OSCAR corpus, a recent Wikipedia dump, various OPUS corpora, and a specialized corpus contributed by Kemal Oflazer \cite{schweter2020berturk}. Additionally, the same team conducted pre-training for DistilBERTurk, ConvBERTurk, and Turkish ELECTRA models. In the literature, there are studies aiming to adapt BERTurk to specific domains through further pre-training on domain-specific corpora. For instance, Bayrak et al.\cite{bayrak} pre-trained BERTurk using a medical-specific corpus, while Türkmen et al. \cite{turkmen2023bioberturk} introduced BioBERTurk, which involved additional pre-training of BERTurk on biomedical and radiology-related corpora. When it comes to the legal domain, to the best of our knowledge, no study involving modifications to the original BERT structure or alterations to subtasks in pre-training (MLM and NSP) for Turkish legal texts has been conducted. Only recently, there has been a limited number of studies in the legal domain, resulting in two models named BERTurk-Legal \cite{ozturk} and HukukBERT \cite{akca}. These models also focus on further pre-training BERTurk on legal-related corpus, and they did not pre-train the BERT model from scratch.

\section{Models}
In contrast to other previous studies, we kept the BERT structure and hyperparameters—such as hidden layers, max-position-embeddings, num-attention-heads, hidden-size (except for the vocabulary size)—unchanged. Our focus was solely on manipulating the pre-training tasks of NSP and MLM. Additionally, we pre-trained our models from scratch without further pre-training on a non-specific pre-trained model. In general, with a focus on NSP and MLM, we pre-trained five proposed models, in addition to the original BERT model (Table 1). For the fine-tuning phase, we have limited the scope of the study to downstream tasks of NER and multi-label classification.
\begin{table}[htbp]
    \raggedright
    \caption{Details of pre-training tasks for each pre-training approach.}
    \centering
    \small
    \begin{tabular}{lccccc}
        \hline
        \textbf{Pre-trained approach} & \textbf{NSP} & \textbf{SOP} & \multicolumn{3}{c}{\textbf{Masking strategy percentages}}\\
        \cline{4-6}     
        & & & \textbf{Masked} & \textbf{Replaced} & \textbf{Unchanged} \\
        \hline
        Original BERT: NSP\_MLM\_80\_10\_10 & * & - & 80 & 10 & 10 \\
        SOP\_MLM\_80\_10\_10 & - & * & 80 & 10 & 10 \\
        MLM\_80\_10\_10 & - & - & 80 & 10 & 10 \\
        MLM\_80\_0\_20 & - & - & 80 & 0 & 20 \\
        MLM\_80\_(10\_TF\_IDF)$^a$\_10 & - & - & 80 & 10 & 10 \\
        MLM\_80\_(20\_TF\_IDF)$^a$\_0 & - & - & 80 & 20 & 0 \\
        \hline
    \end{tabular}
    \smallskip
    \begin{flushleft}
        Note: \\
        $^a$ The MLM task replaces the MLM selected token using tokens with a high TF-IDF score instead of utilizing the tokenizer.
    \end{flushleft}
    \label{tab:table1}
\end{table}
As shown in Table ~\ref{tab:table1}, the original BERT model is labeled as "NSP\_MLM\_80\_10\_10" and involves two subtasks: NSP and MLM\_80\_10\_10. In the context of MLM, the notation (80\_10\_10) indicates that among the randomly chosen 15\% of tokens (referred to as MLM selected tokens), 80\% are masked, 10\% are replaced by a randomly selected token from the tokenizer vocabulary, and the remaining 10\% remain unchanged \cite{devlin2018bert}.
Regarding modifications to the NSP task, as highlighted in section 2, several researchers have suggested that the NSP task might not be essential. They proposed that altering or removing it could be a preferable approach \cite{lan2019albert,wang2019structbert,liu2019roberta,yang2019xlnet,milosheski2020lambert,iter2020pretraining,bai2020segabert,clark2020electra,lample2019crosslingual,joshi2020spanbert}. Their findings resulted in significant deviations in BERT configuration or the creation of completely new models. However, in contrast to this research, we have chosen to maintain the unaltered BERT-based configuration and subsequently propose replacing NSP with SOP since SOP provides a more comprehensive alternative to NSP. The model proposed is identified as "SOP\_MLM\_80\_10\_10". Additionally, we acknowledge the potential presence of noise and lengthy sentences in legal texts, often exceeding 512 tokens. Considering this, it is reasonable that NSP or SOP might not yield significant performance improvements in the pre-training task. Therefore, we recommend retaining MLM and removing NSP, identified as "MLM\_80\_10\_10".
Regarding modifications to MLM, in the original MLM \cite{devlin2018bert}, the "10" in "MLM\_80\_\textbf{10}\_10" rule indicates that 10\% of the MLM-selected tokens are replaced by random tokens from the tokenizer's vocabulary list. We propose an innovative approach: replacing these tokens with high TF-IDF valued tokens, recognized as keywords across all documents (in a keywords list). This modified model is named "MLM\_80\_(10\_TF\_IDF)\_10". Furthermore, we experimented with different masking strategies in MLM and introduced two alternative approaches: "MLM\_80\_0\_20" involves no replacement of random tokens, while "MLM\_80\_(20\_TF\_IDF)\_0" replaces 20\% of MLM-selected tokens with random tokens from a list of tokens with high TF-IDF scores.
Given our limited corpus size and the insufficient data available for training the tokenizer, we made the decision to utilize a BERTurk tokenizer (dbmdz/bert-base-turkish-cased), with a vocabulary size of 32,000 \cite{schweter2020berturk}. For BertConfig, we selected the BERT\_base structure with 12 hidden layers, maintaining the default settings of BERT\_base except for the vocab\_size, which we set to 32,000 to align with the BERTurk tokenizer. Models were pre-trained with a single epoch, and due to GPU limitations, the batch size was considered as 16. The most important BertConfig hyperparameters are summarized in Table ~\ref{tab:table2}.
\begin{table}[htbp]
    \raggedright 
    \caption{Key BERTConfig parameters for pre-training the models.}
    \centering
    \begin{tabular}{lc}
        \toprule
        \textbf{Parameter} & \textbf{Value} \\
        \hline
        Vocab\_size & 32,000 \\
        Max\_position\_embeddings & 512 \\
        Num\_attention\_heads & 12 \\
        Num\_hidden\_layers & 12 \\
        Hidden\_size & 768 \\
    \bottomrule
    \end{tabular}
    \label{tab:table2}

\end{table}
Hyperparameters for the fine-tuning process in both NER and multi-label text classification are detailed in Table ~\ref{tab:table3}. In both methods, the models undergo fine-tuning for a single epoch, employing a train\_batch\_size of 128.
We employed the NVIDIA Quadro RTX 8000 GPU (VCQRTX8000-PB, 48GB GDDR6 ECC, Professional 3D) to train our language models in both pre-training and fine-tuning phases.
\begin{table}[htbp]
    \raggedright
    \caption{Fine-tuning configuration for NER and multi-label text classification.}
    \centering
    \begin{tabular}{lc}
        \toprule
        \textbf{Parameter} & \textbf{Value/Setting} \\
        \hline
        Train\_batch\_size & 128 \\
        Train\_epochs & 1 \\
        Optimizer & AdamW \\
        Learning\_rate & 4e-5 \\
        Scheduler & Linear\_schedule\_with\_warmup \\
    \bottomrule
    \end{tabular}
    \label{tab:table3}
\end{table}

\section{Dataset Collection }
This section concentrates on providing a comprehensive explanation of the corpus and annotated data used for pre-training and fine-tuning phases.
\subsection{Corpus in the Legal Turkish Domain}
Due to data sensitivity in the legal Turkish domain, especially with personal information, sharing such data is impractical for companies. To overcome this, we utilized thesis texts on legal topics from YÖKTEZ, downloaded in PDF format from the YÖKTEZ website \footnote{\href{https://tez.yok.gov.tr/UlusalTezMerkezi/tarama.jsp}{https://tez.yok.gov.tr/UlusalTezMerkezi/tarama.jsp} }. The extracted text from these files serves as the legal corpus for pre-training. Subsequently, each file's text is divided into sentences using Spacy's sentencizer. The YÖKTEZ corpus comprises 14,815 files, totaling 4.35 GB. During preprocessing, we excluded files that were 1 KB in size, containing only the first page of the thesis with student and supervisor names. Theses with multiple languages and files containing symbols and non-Turkish characters were also identified and excluded.
After cleaning the data, the corpus was reduced to 3.5 GB with 13,094 files. Due to resource limitations, we initially identified the best models using a randomly selected 50 MB subset, including 124 files and 347,147 sentences (refer to Table ~\ref{tab:table4} for further details). Following the identification of the best models, these models were pre-trained on various corpus sizes: 100 MB (235 files, 692,164 sentences), 500 MB (1,260 files, 3,269,388 sentences), 1 GB (2,449 files, 6,548,666 sentences), and 2 GB (4,821 files, 12,854,227 sentences).
\begin{table}[htbp]
    \raggedright
    \caption{Corpus details for the pre-training stage.}
    \centering
    \begin{tabular}{lc}
        \toprule
        \textbf{Attribute} & \textbf{Value} \\
        \hline
        Size of the corpus & 50 MB \\
        Number of files in the corpus & 124 \\
        Total sentences in the corpus & 347,147 \\
        Total number of tokens & 11,543,970 \\
        Total number of unique tokens & 27,280 \\
        Total number of words & 6,465,260 \\
        Total number of unique words & 399,104 \\
    \bottomrule
    \end{tabular}
    \label{tab:table4}
\end{table}
\subsection{Human-annotated Datasets }
NewMind\href{https://www.newmind.ai/en/ }{\footnote{\href{https://tez.yok.gov.tr/UlusalTezMerkezi/tarama.jsp}{https://www.newmind.ai/en/} }}, a legal tech company in Istanbul, Turkey, supplied human-annotated datasets. The datasets, created by a skilled team of legal experts and law students, support both NER and multi-label text classification tasks in this project.
\subsubsection{Annotated data in NER task}
When utilizing machine learning for NER tasks, annotated data is often structured using BIO (Beginning, Inside, Outside) tags. This format employs 'B' for entity beginnings, 'I' for inside entities, and 'O' for non-entities. An alternative method, BIESO (Beginning, Intermediate, End, Single word entity, Outside) provides more detailed tagging, adding distinctions for 'end' and 'intermediate' tokens, as well as for single-word versus multi-word entities \cite{tang2013recognizing}. In this study, the chosen format is BIO, demonstrated in Table ~\ref{tab:table5} using the sentence: "Burada, tebligat kanunu ile vuk düzenlemesi ayrımına dikkat etmek gerekir." (English version: "Here, it is important to pay attention to the distinction between the notification law and the tax procedure law."), wherein "tebligat kanunu/ notification law" and "vuk/ tax" are labeled as "law". 
\begin{table}[htbp]
    \raggedright
    \caption{Example of a sentence in BIO format used for NER tasks.}
    \centering
    \begin{tabular}{ll}
        \toprule
        \textbf{Words} & \textbf{Labels} \\
        \hline
        Burada & O \\
        , & O \\
        tebligat & B-Law \\
        kanunu & I-Law \\
        ile & O \\
        vuk & B-Law \\
        düzenlemesi & O \\
        dikkat & O \\
        etmek & O \\
        gerekir & O \\
        . & O \\
    \bottomrule
    \end{tabular}
    \label{tab:table5}

\end{table}
The annotated data consists of 614,541 sentences, occupying a total of 305 MB. Additionally, the count of labelled tokens reaches 20,220,789. This dataset includes 11 general labels covering diverse categories such as 'Person', 'Law', 'Publication', 'Government', 'Court', 'Date', 'Money', 'Location', 'Corporation', 'Other Organization', and 'Project'.

\subsubsection{Annotated data in multi-label text classification task}
The provided annotated dataset for the multi-label text classification task consists of 175,279 sentences, collectively occupying 63 MB of storage. Table ~\ref{tab:table6} showcases two sentence examples along with their multi-label assignments. The columns in the table represent various labels, where a value of "1" signifies the presence of that label for the sentence, and "0" indicates the absence of that label. Additionally, to maintain data sensitivity, certain information like organization names is anonymized as "XXX". 
\begin{table}[htbp]
    \raggedright
    \caption{An example of multi-label text classification.}
    \centering
    \begin{tabular}{p{6cm}cccc}
        \toprule
        \textbf{Sentence} & \textbf{Heading} & \textbf{Contract} & \textbf{Price} & \textbf{Taxes} \\
        \hline
        MADDE 3 - FİYAT SİTE YÖNETİMİ, işbu Sözleşme ile, XXX ’ ın SİTE YÖNETİMİ ’ ne vereceği okuma ve faturalandırma işlemi için XXX’ a aşağıda belirtilen bedelde ödeme yapacaktır. 
        
        [English version: "ARTICLE 3 - SITE MANAGEMENT FEE, pursuant to this Agreement, XXX will make a payment to XXX at the below-mentioned rate for the reading and invoicing process that XXX will provide for SITE MANAGEMENT."] & 1 & 1 & 0 & 0 \\
        \hline
        XXX tarafından bireysel ayrıntılı gider paylaşım bildirgesinin ( tüm sakinlerin bireysel bildirgesi ) e - posta ile SİTE YÖNETİMİ ’ ne gönderilmesi aylık, daire başına 5.00 TL + KDV üzerinden hesaplanacaktır. 
        
        [English version: "Sending individual detailed expense sharing statements (individual statements of all residents) to SITE MANAGEMENT via e-mail will be calculated on a monthly basis at 5.00 TL + VAT per apartment." ] & 0 & 1 & 1 & 1 \\
    \bottomrule
    \end{tabular}
    \label{tab:table6}

\end{table}
The dataset encompasses 31 labels, which include categories such as 'Heading', 'Penalty', 'Transfer\_and\_assignment', 'Termination', 'Confidentiality', 'Non liability clause', 'Indemnity', 'Waiver', 'Force majeure', 'Change of control', 'Protection of personal data', and others.

\section{Models Performance Evaluation  }
Considering the time-consuming and expensive nature of training the five proposed models plus the original BERT model in both the pre-training and fine-tuning phases, we have adopted a two-step evaluation approach. In the first step, using a 50MB corpus, we aim to identify the best model for NER and multi-label text classification. In the second step, only the best-performing models will undergo pre-training on a larger corpus, and their performance will be compared with BERTurk.

\subsection{Evaluation with 50MB corpus }
\begin{figure}
    \raggedright
    \includegraphics[width=1\linewidth]{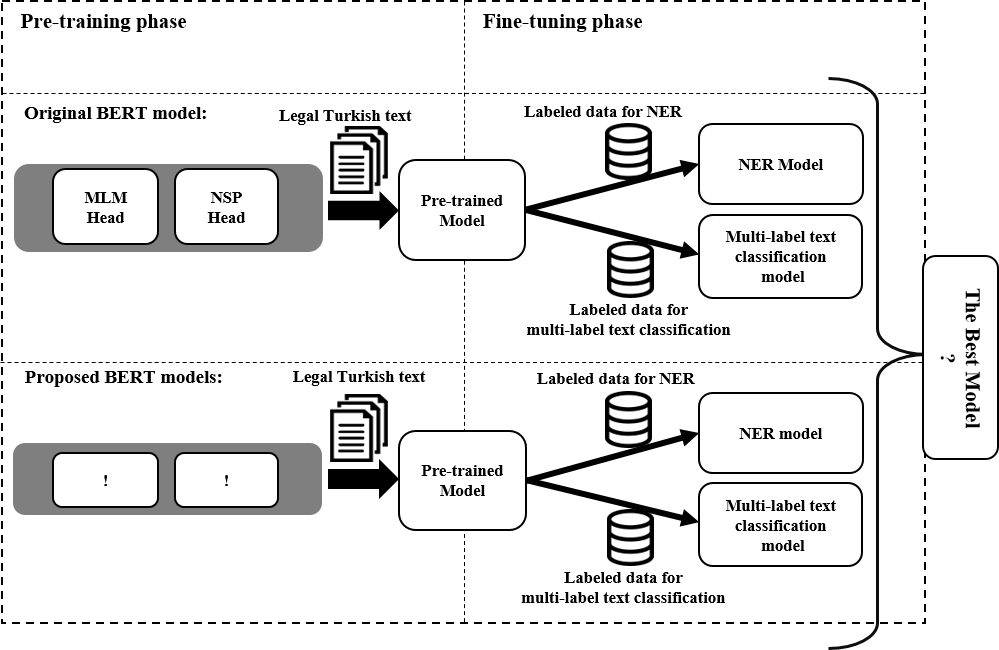}
    \caption{Evaluation process for pre-trained models.}
    \centering
    \label{fig:fig1}
\end{figure}

As illustrated in Figure ~\ref{fig:fig1}, each proposed model undergoes pre-training on a 50 MB Turkish legal corpus. Subsequently, each model is fine-tuned for two downstream tasks: NER and multi-label text classification. Following the fine-tuning process, the performance of all models will be assessed, and the best pre-trained models will be identified.
To fine-tune models, the Hold-out method is used to partition the datasets into training, validation, and test sets. Specifically, 80 percent of the datasets are allocated for training, while the remaining 10 percent are reserved for validation and another 10 percent for testing. Table ~\ref{tab:table7} provides information on the number of sentences in both annotated data (NER and multi-label text classification) during the fine-tuning stage. Moreover, Precision, Recall, and F-measure are employed to assess the fine-tuned model's performance. 
\begin{table}[htbp]
    \raggedright
    \caption{Distribution of sentences in annotated data for NER and multi-label text classification during fine-tuning.}
    \centering
    \begin{tabular}{lccc}
        \toprule
      \textbf{Dataset} & \textbf{Split} & \textbf{NER data} & \textbf{Multi-label text classification data} \\
        \hline
        Train set & 0.8 & 501,661 & 137,955 \\
        Validation set & 0.1 & 56,440 & 18,662 \\
        Test set & 0.1 & 56,440 & 18,662 \\
        Total sentences & 1.0 & 614,541 & 175,279 \\
    \bottomrule
    \end{tabular}
    \label{tab:table7}
\end{table}
\subsection{Evaluation on larger corpora}
After identifying the best models on a 50MB corpus, they will be further evaluated on larger corpora (100MB, 500MB, 1GB, and 2GB). The performance of these models will be compared with BERTurk "dbmdz/bert-base-turkish-cased", which uses BERT base structures and is pre-trained on a 35GB corpus.

\section{Results and discussion}
Table ~\ref{tab:table8} summarizes the fine-tuning results of models trained from scratch. This encompasses a total of 18 training rounds: 6 pre-training rounds on the 50MB corpus, followed by 12 rounds of fine-tuning for NER and multi-label text classification. Further details are discussed in the following sections.

\begin{table}[htbp]
    \raggedright
    \caption{Summary of fine-tuning results for models pre-trained from scratch.}
    \centering
    \small
    \begin{tabular}{p{3cm}ccc|ccc}
        \toprule
        \textbf{Pre-trained models} & \multicolumn{3}{c|}{\textbf{Multi-label Text classification}} & \multicolumn{3}{c}{\textbf{NER}} \\
        \cline{2-7}     
        & \textbf{Precision} & \textbf{Recall} & \textbf{F-measure} & \textbf{Precision} & \textbf{Recall} & \textbf{F-measure} \\
        \hline     
        \textbf{Original BERT:} NSP\_MLM\_80\_10\_10 & 84.36 & 48.79 & 61.82 & 50.77 & 73.53 & 60.07 \\
        SOP\_MLM\_80\_10\_10 & 84.93 & 54.71 & 66.55 & 51.22 & 73.93 & 60.51 \\
        MLM\_80\_10\_10 & 86.18 & \textbf{61.88} & 72.03 & 70.33 & 82.33 & 75.86 \\
        MLM\_80\_0\_20 & 86.31 & 60.05 & 70.82 & \textbf{71.85} & \textbf{83.47} & \textbf{77.23} \\
        MLM\_80\_(10\_TF\_IDF)\_10 & 86.59 & 61.75 & 72.09 & 68.73 & 81.62 & 74.62 \\
        MLM\_80\_(20\_TF\_IDF)\_0 & \textbf{86.61} & 61.83 & \textbf{72.15} & 67.53 & 81.15 & 73.72 \\
    \bottomrule
    \end{tabular}
    \label{tab:table8}
    \smallskip
\end{table}

\subsection{Results of replacing NSP with SOP  }
When comparing the SOP-based model (SOP\_MLM\_80\_10\_10) with the original BERT model (NSP\_MLM\_80\_10\_10), a positive influence of this substitution on both NER and multi-label text classification tasks is evident. In multi-label text classification, the f-measure experiences a notable improvement, surging from 61.82 in the original BERT to 66.55 in the SOP-based model. This improvement underscores the effectiveness of utilizing SOP in enhancing the model's ability to classify sentences into multiple categories. In NER, the f-measure exhibits a slight yet meaningful advancement, considering the small corpus size. The result signifies the positive impact of SOP on the model's capacity to recognize named entities within text.
\subsection{Results of retaining MLM and removing NSP  }
Compared to the original BERT model (NSP\_MLM\_80\_10\_10) and the model without the NSP component (MLM\_80\_10\_10), significant improvements are seen in both multi-label text classification and NER tasks. The f-measure notably increases from 61.82 to 72.03 in multi-label text classification and from 60.07 to 75.86 in NER.
Additionally, when comparing the results of replacing NSP with SOP on model performance (SOP\_MLM\_80\_10\_10), it becomes evident that, in general, retaining MLM and removing NSP has shown better performance compared to the alternative of replacing NSP with SOP. The result indicates that in the legal domain, where there are long sentences that can include noises, both NSP and SOP, focusing on sentence relations can make the model confusing and reduce performance.
\subsection{Results of combining MLM with TF-IDF and diverse masking strategies  }
\begin{figure}
    \raggedright
    \includegraphics[width=1\linewidth]{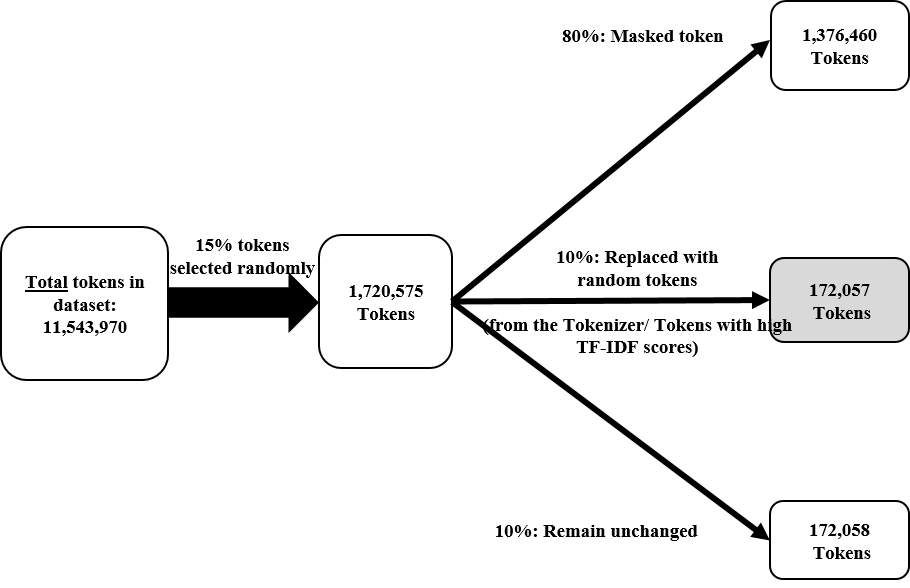}
    \caption{Token selection and masking distribution in MLM processes. }
    \centering
    \label{fig:fig2}
\end{figure}

Based on the results presented in previous sections, we have chosen to narrow our focus to MLM. We compare the original MLM\_80\_10\_10 with a model that combines MLM with TF-IDF using different masking strategies. 
During the implementation of code and dataset creation for both "MLM\_80\_10\_10" and "MLM\_80\_(10\_TF\_IDF)\_10", Figure ~\ref{fig:fig2} shows that 15\% of all tokens (1,720,575 tokens) were selected in the MLM process. Among these, 80\% (1,376,460 tokens) were masked, 10\% (172,057 tokens) were replaced by random tokens from the tokenizer or tokens with high TF-IDF scores, and the remaining 10\% (172,058 tokens) remained unchanged. 

We further analyzed the 172,057 tokens replaced by random tokens from the tokenizer or tokens with high TF\_IDF scores. Table ~\ref{tab:table9} compares the first 100 tokens from the tokenizer and the high TF\_IDF score token list. Tokens from the TF\_IDF list are more contextually relevant, contrasting with tokenizer tokens that may lack context. For example, tokenizer tokens include personal names (e.g., "Hakan"), country names (e.g., "İran"), numbers, symbols (e.g., "\#\#"), and abbreviations (e.g., "CR"). In contrast, high TF\_IDF tokens are highly relevant to legal texts, such as "yeterlik/ sufficiency", "hususunda/ about the issue", "komisyon/ commission", "kanunu/ law" , "sonuçların/ result", "müdahalede/ intervention", "hükümleri/ provisions".

\begin{table}[htbp]
    \raggedright
    \caption{Comparison of randomly chosen tokens in “MLM\_80\_10\_10” and MLM\_80\_(10\_TF\_IDF)\_10”.}
    \centering
    \small
    \begin{tabular}{p{6cm}|p{6cm}}
        \hline
        \textbf{Selected tokens from Tokenizer In MLM\_80\_10\_10} & \textbf{Selected tokens from TF\_IDF list In MLM\_80\_(10\_TF\_IDF)\_10} \\
        \hline
        bany, emisyon, larından, katil, ABD, akraba, Kayıp, departman, teki, Şili, neleri, memiştim, çıkmadan, kapsayacak, maları, sezonun, kalbinde, \$, hücreleri, Antal, 550, Kes, erecek, mişler, saldırısı, 93, vatandaşları, tezah, üssü, mesiyle, ting, sıkıntılar, nemli, CR, spekülasyon, gör, Hakan, Yugoslavya, 46, kuruluşunun, arkan, NO, BA, malzemeleri, star, eğiyle, pank, roid, ic, aksi, yerinin, MAZ, hun, ecilik, hatırlar, Sonu, önceki, ĝ, bilim, Ø, lendiğini, Harvard, risk, var, Kle, DU, sağlanmıştır, memnuniyeti, tığımız, İran, layacağını, sektörünün, G, dizinin, @, yükümlül, Balkan, şok, komit, unu, Kağıt, Süper, beklet, \%, milliyet, Edirne, yed, **, \#\#törler, mıştık, (, miyim, ây, verilerini, \#\#, keyifle, 000, Fo, DO, dağıt, \_, imzalı, , ödev, \#\#Ä±, \#\#ılmadı, ödemeleri, !, sahiplerinin, eksiklik, eð, nü, yürüyüşe, SAT, Anal, ș, yenik, \#\#ılmaktadır, zenci, fah, Dergi, Finlandiya, baz, … 
        &incelemesi, değişebilir, gömülü, alındıktan, bütçeye, savaşan, hepimizin, yeterlik, üreticiler, duyunca, ilah, kaynat, komisyon, bazında, etmişlerdir, dola, sağlarken, doğrulama, şarttır, doğruluk, zenginlik, temizliği, irtibata, kanunu, projesinin, sonuçların, huzur, işletmeler, sıf, müdahalede, çıkarıyor, sunmaktayız, sük, hükümleri, taşıyan, odayı, opt, enflasyonun, gösteriş, teçhiz, hususunda, adedi, istedim, öğretmenlere, bütçeye, yükselmesi, kadro, işlemleri, fiyat, anlaşma, anlayışını, siyasal, düşmesine, maddelerin, öreninde, seansta, lakin, arzu, tertip, getirdik, şart, karakterler, şeffaf, kaybedecek, tutarak, devlete, hazırlandı, ettiklerini, borçlarını, toplantısı, sağlama, , hizmetler katılmış, seçim, tüzel, belediyeler, verilen, dileriz, karakterini, zorunda, iddi, geçirir, karşıyay, zorundayız, çöküş, , tahvil, odayı, kaydet, yetin, önleyici, değişti, halle, mesel, zarar, yaşar, istihdam, ihti, … \\
    \bottomrule
    \end{tabular}
    \label{tab:table9}
    \smallskip
\end{table}
When examining the results of the trained models presented in Table 8, which demonstrate the effect of replacing random tokens from the tokenizer or a token list with high TF\_IDF score, in multi-label text classification, introducing more random tokens generally enhances performance compared to the original MLM\_80\_10\_10. Furthermore, utilizing tokens from a token list with high TF-IDF scores, instead of selecting tokens from the tokenizer, also yields positive results. In "MLM\_80\_0\_20", where no random tokens are added, the f-measure decreases. In "MLM\_80\_(10\_TF\_IDF)\_10", where random tokens are chosen based on high TF-IDF scores, there is a slight improvement in f-measure. Additionally, in "MLM\_80\_(20\_TF\_IDF)\_0" with more random tokens (20\%), further improvement is shown.
In NER, contrary to multi-label text classification, the addition of random tokens negatively affects performance compared to the original MLM\_80\_10\_10. In “MLM\_80\_0\_20”, where no random tokens are added, the f-measure increases. In “MLM\_80\_(10\_TF\_IDF)\_10”, where random tokens are selected based on high TF-IDF scores, there is a slight decrease in f-measure. In “MLM\_80\_(20\_TF\_IDF)\_0”, this negative impact of adding random tokens persists, with a further reduction in f-measure. This result underscores that for NER, which focuses on each token individually, introducing random tokens introduces too much distraction.
\subsection{Results of models trained on a larger corpus  }
We pre-trained our best models for both multi-label text classification "MLM\_80\_(20\_TF\_IDF)\_0" and the NER task "MLM\_80\_0\_20" on a larger corpus. To achieve this, we augmented our initial 50 MB corpus by incorporating additional random documents, expanding the corpus size to 100 MB, 500 MB, 1 GB, and 2 GB. Simultaneously, we compared our domain-specific pre-trained models with Schweter's BERTurk, which was pre-trained on a 35 GB non-specific corpus containing 4,404,976,662 tokens. Subsequently, we fine-tuned these models on the same annotated data for both NER and multi-label text classification tasks. Table ~\ref{tab:table10} presents the results.
\begin{table}[htbp]
    \raggedright
    \centering

    \caption{Performance comparison between our modified models and BERTurk. }
    \small
    \begin{tabular}{p{1.2cm}p{3.5cm}ccc|ccc}
        \toprule
        \multirow{2}{*}{\textbf{Corpus}} & \multirow{2}{*}{\textbf{Pre-trained model}}  & \multicolumn{3}{c|}{\textbf{Multi-label classification}} & \multicolumn{3}{c}{\textbf{NER}} \\
        \cline{3-8}     
         \textbf{size}& & \textbf{Precision} & \textbf{Recall} & \textbf{F-measure} & \textbf{Precision} & \textbf{Recall} & \textbf{F-measure} \\
        \hline     
         35 GB& \textbf{BERTurk:} NSP\_MLM\_80\_10\_10  & 88.88 & \textbf{72.59} & \textbf{79.91} & 90.75 & \textbf{94.25} & 92.46 \\
        \hline     
         50 MB & MLM\_80\_(20\_TF\_IDF)\_0 & 86.61 & 61.83 & 72.15 & - & - & - \\
          & MLM\_80\_0\_20 & - & - & - & 71.85 & 83.47 & 77.23 \\
        \hline     
         100 MB & MLM\_80\_(20\_TF\_IDF)\_0 & 86.76 & 61.88 & 72.24 & - & - & - \\
          & MLM\_80\_0\_20 & - & - & - & 81.52 & 88.06 & 84.67 \\
        \hline     
        500 MB & MLM\_80\_(20\_TF\_IDF)\_0 & 88.37 & 68.93 & 77.45 & - & - & - \\
         & MLM\_80\_0\_20 & - & - & - & 88.02 & 92.94 & 90.41 \\
        \hline     
         1 GB & MLM\_80\_(20\_TF\_IDF)\_0 & 88.62 & 69.45 & 77.87 & - & - & - \\
         & MLM\_80\_0\_20 & - & - & - & 89.09 & 93.39 & 91.19 \\
        \hline

        2 GB & MLM\_80\_(20\_TF\_IDF)\_0 & \textbf{88.91} & 70.2 & 78.45 & - & - & - \\
         & MLM\_80\_0\_20 & - & - & - & \textbf{90.89} & 94.12 & \textbf{92.48} \\
    \bottomrule
    \end{tabular}
    \label{tab:table10}
    \smallskip
\end{table}

As seen in Table 10, expanding the corpus from 50 MB to 2 GB significantly improved models' performance. The f-measure for the NER task increased from 77.23 to 92.48, and for multi-label classification, it rose from 72.15 to 78.45. This indicates that an even larger corpus could bring further enhancements.

In comparison with the results from BERTurk, our models, pre-trained on a 2 GB corpus for the NER task, demonstrate superior precision and f-measure. In multi-label text classification, our models surpass in precision, despite BERTurk being pre-trained on a 35 GB corpus, which is more than 17 times larger than our corpus. However, a hypothesis is considered that with an even larger corpus related to legal text, our models may achieve even better performance than BERTurk.

\section{Future work   }
This study, like all other studies, was conducted with certain limitations. In future studies, these limitations can be addressed to explore different research possibilities. Regarding the corpus used in the pre-training phase, due to the sensitive nature of legal data, we used text pre-training extracted from publicly available legal domain theses. However, we plan to use texts from legal professionals' files to create a more representative corpus. Additionally, our focus was solely on legal Turkish text, and we plan to explore the use of various fields in Turkish and also a multilingual corpus to evaluate the potential for generalizing results. From a technical perspective, we exclusively focused on two downstream tasks, namely NER and multi-label text classification. Therefore, we want to extend our analysis to include other NLP downstream tasks, such as question answering and binary label text classification. Furthermore, we only trained all models, both in the pre-training stage and the fine-tuning stage, for a single epoch. It is planned to conduct longer training sessions with multiple epochs to potentially achieve better results. Additionally, in our effort to enhance the MLM subtask by combining TF-IDF, our focus was on tokens rather than whole words. It is possible to adapt this approach to work at the whole-word level and investigate if it can yield improved results. Moreover, we showed that using a larger corpus during the pre-training phase can lead to improved model performance. Therefore, it is planned to work with a larger corpus to achieve a more robust generalization of results.  
\section{Conclusion    }
While Google's BERT has demonstrated remarkable performance, there is an opportunity for further enhancements, particularly when tailoring these models for specific domains and non-English contexts. After reviewing other studies within the legal Turkish domain, we found no prior research focusing on enhancing BERT's performance within this field. Therefore, in this paper, we aimed to improve the BERT model's performance within the legal Turkish domain.
 We undertook the task of pre-training specialized versions of the BERT model from scratch using a legal Turkish corpus extracted from texts related to legal topics in thesis texts. Unlike other studies, we kept the BERT-base config unchanged, except for the vocabulary size. We pursued three main ideas to enhance BERT's performance, which included replacing NSP with SOP, removing NSP, and combining MLM with TF-IDF, while also applying different masking strategies within the original MLM\_80\_10\_10. Subsequently, we fine-tuned these customized models for NER and multi-label text classification.

We demonstrated that replacing NSP with SOP generally has a positive impact on the BERT model. However, the removal of NSP/SOP has a much more significant influence on the outcomes due to the presence of noise and lengthy sentences in legal texts. Subsequently, we focused solely on MLM, and after applying different masking strategies in MLM, it was shown that, in multi-label text classification, the substitution of random tokens demonstrates a favorable effect. The best-performing model in multi-label text classification is MLM\_80\_(20\_TF\_IDF)\_0, where 20\% of MLM-selected tokens are replaced with random tokens from those with high TF-IDF values, which have significant meaning in context and are more related to the context compared to random tokens from the tokenizer. In contrast, in the NER task, the addition of random tokens negatively impacts the results. The most effective model for this task is MLM\_80\_0\_20, where only the MLM is employed, and 80\% of MLM-selected tokens are masked without any random token replacements. In comparison with the original BERT model, our proposed models achieved higher f-measures of 72.15\% and 77.23\% in multi-label text classification and NER, respectively.
Additionally, we pre-trained our best models on a larger corpus (100 MB, 500 MB, 1 GB and 2 GB). Generally, we showed that with an increase in the corpus size, the performance of models improves. When comparing with BERTurk results, which was pre-trained on 35 GB non-specific Turkish corpus, our models pre-trained on a 2 GB Turkish legal-related corpus outperform in f-measure and precision in NER and in precision in multi-label classification tasks.

\section*{Disclosure statement}
No potential conflict of interest was reported by the authors.

\section*{Acknowledgments}
This paper is the result of the Ph.D. thesis titled “Improving the Performance of NLP Tasks in Legal Tech” with the reference number 10609878, conducted in collaboration between the Informatics Department, Istanbul University, and NewMind Company. 

\section*{Notes on contributors}
\textbf{Farnaz Zeidi} is an expert in Machine Learning and Natural Language Processing (NLP). She began her Ph.D. in 2018 in the field of Informatics at Istanbul University and, in 2022, continued as a postdoctoral researcher specializing in NLP at the Paul Ehrlich Institute in Germany. She has developed and implemented state-of-the-art NLP and machine learning solutions for extracting, analyzing, summarizing, and generating information from texts in German, English, and Turkish across various domains, including legal and medical.

\textbf{Mehmet Fatih Amasyali } received the MS degree from Yildiz Technical University, Turkey in 2003, and the PhD degree from the same university in 2008. He did his postdoc study at Purdue University, West Lafayette School of Electrical and Computer Engineering, USA. Dr. Amasyali is currently a Professor at the Department of Computer Engineering, Yildiz Technical University, Turkey. His interests include machine learning, natural language processing and autonomous robotics. He has published several scientific papers. 

\textbf{Çiğdem Erol} is a Professor in the Computer Sciences Department (Faculty of Science) in Istanbul University, Turkey. She received her Ph.D. degree from Istanbul University Institute of Science in 2010. Her research interests are (Bio)Informatics, Computer Sciences, Information Technologies, Data Mining, Machine Learning.
\bibliographystyle{unsrt}  
\bibliography{references}

\end{document}